\documentclass[letterpaper]{article} 
\usepackage{aaai24}  
\usepackage{times}  
\usepackage{helvet}  
\usepackage{courier}  
\usepackage[hyphens]{url}  
\usepackage{graphicx} 
\usepackage{natbib}  
\usepackage{caption} 
\usepackage{booktabs}       
\usepackage{amsfonts}       
\usepackage{subcaption}
\usepackage{amsthm}
\newtheorem{theorem}{Theorem}
\newtheorem*{proof*}{Proof} 
\usepackage{here}
\usepackage{multirow}
\usepackage{multicol}
\usepackage{graphicx}
\usepackage{multirow}

\usepackage{amsmath}

\usepackage{algorithm}
\usepackage{algorithmic}

\usepackage{newfloat}
\usepackage{listings}
\DeclareCaptionStyle{ruled}{labelfont=normalfont,labelsep=colon,strut=off} 
\floatstyle{ruled}
\newfloat{listing}{tb}{lst}{}
\floatname{listing}{Listing}
\title{FairSIN: Achieving Fairness in Graph Neural Networks \protect\\through Sensitive Information Neutralization}
\author {
    Cheng Yang\textsuperscript{\rm 1},
    Jixi Liu\textsuperscript{\rm 1},
    Yunhe Yan\textsuperscript{\rm 1},
    Chuan Shi\textsuperscript{\rm 1}\thanks{Corresponding author.}
}
\affiliations {
    \textsuperscript{\rm 1}Beijing University of Posts and Telecommunications, China \\
    \{yangcheng, landon, yur1, shichuan\}@bupt.edu.cn
}

\usepackage{bibentry}

\begin{document}

\maketitle

\begin{abstract}
Despite the remarkable success of graph neural networks (GNNs) in modeling graph-structured data, like other machine learning models, GNNs are also susceptible to making biased predictions based on sensitive attributes, such as race and gender. For fairness consideration, recent state-of-the-art (SOTA) methods propose to filter out sensitive information from inputs or representations, \textit{e.g.,} edge dropping or feature masking. However, we argue that such \textit{filtering-based} strategies may also filter out some non-sensitive feature information, leading to a sub-optimal trade-off between predictive performance and fairness.
To address this issue, we  unveil an innovative \textit{neutralization-based} paradigm, where additional \textit{Fairness-facilitating Features (F3)} are incorporated into node features or representations before message passing. The \textit{F3} are expected to statistically neutralize the sensitive bias in node representations and provide additional non-sensitive information. We also provide theoretical explanations for our rationale, concluding that \textit{F3} can be realized by emphasizing the features of each node's heterogeneous neighbors (neighbors with different sensitive attributes). We name our method as FairSIN, and present three implementation variants from both data-centric and model-centric perspectives.
Experimental results on five benchmark datasets with three different GNN backbones show that FairSIN significantly improves fairness metrics while maintaining high prediction accuracies. Codes and appendix can be found at \url{https://github.com/BUPT-GAMMA/FairSIN}.
\end{abstract}

\section{Introduction}

Graph neural networks (GNNs) have shown their strong ability in modeling structured data, and are widely used in a variety of applications, \textit{e.g.,}  e-commerce~\cite{li2020hierarchical,niu2020dual} and drug discovery \cite{xiong2021graph,bongini2021molecular}. Nevertheless, recent studies~\citep{nifty,fairgnn,chen2021structured,shumovskaia2021linking,li2021dyadic} show that the predictions of GNNs could be biased towards some demographic groups defined by sensitive attributes, \textit{e.g.,} race~\cite{nifty} and gender~\cite{lambrecht2019algorithmic}. In decision-critical applications such as Credit evaluation~\cite{yeh2009comparisons}, the discriminatory predictions made by GNNs may bring about severe societal concerns.

\begin{figure}[t]
  \centering
    \subfloat[\textbf{Pokec-n}]{
    \includegraphics[scale=0.2]{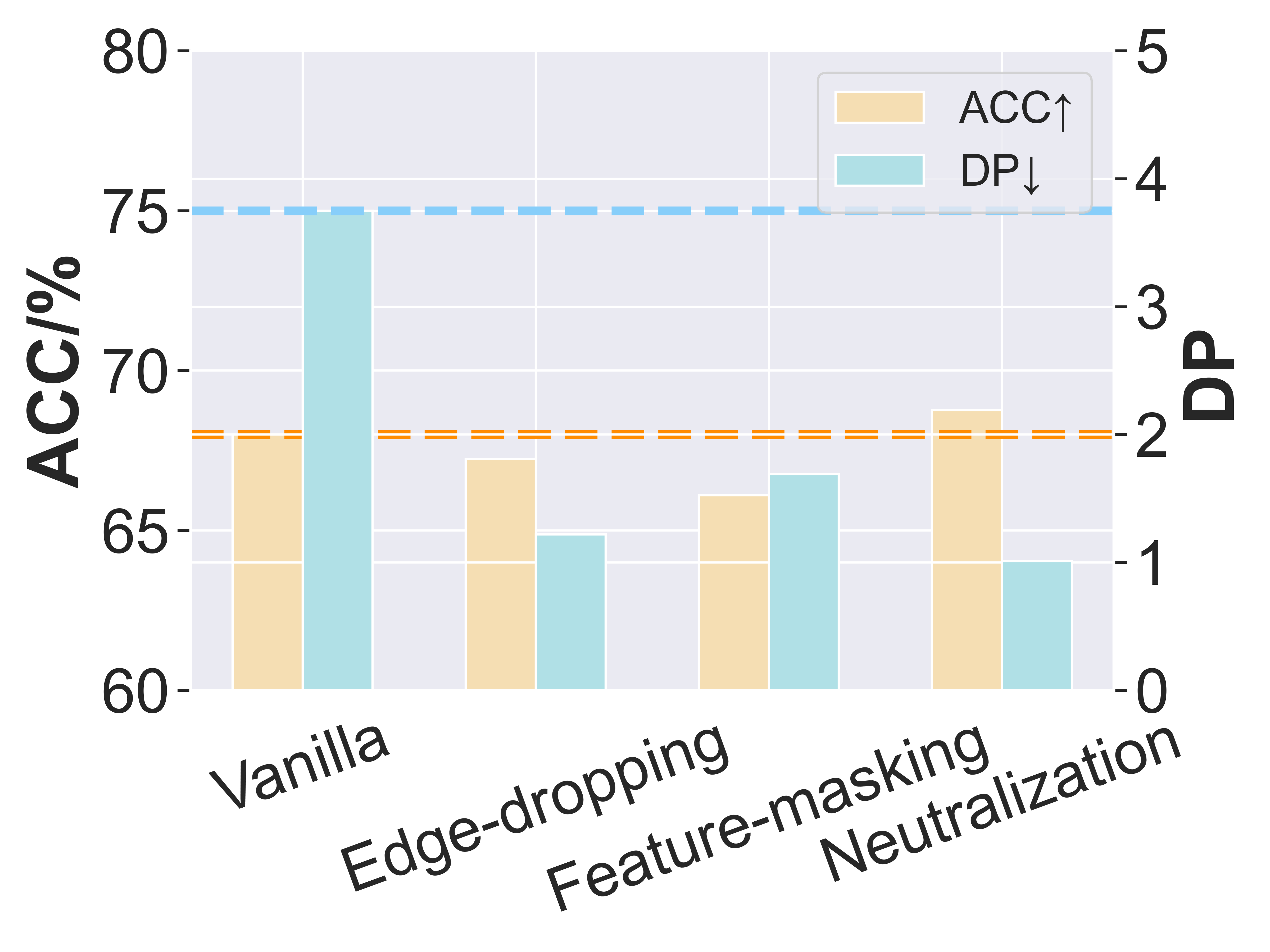}
  }
  \subfloat[\textbf{Pokec-z}]{
    \includegraphics[scale=0.2]{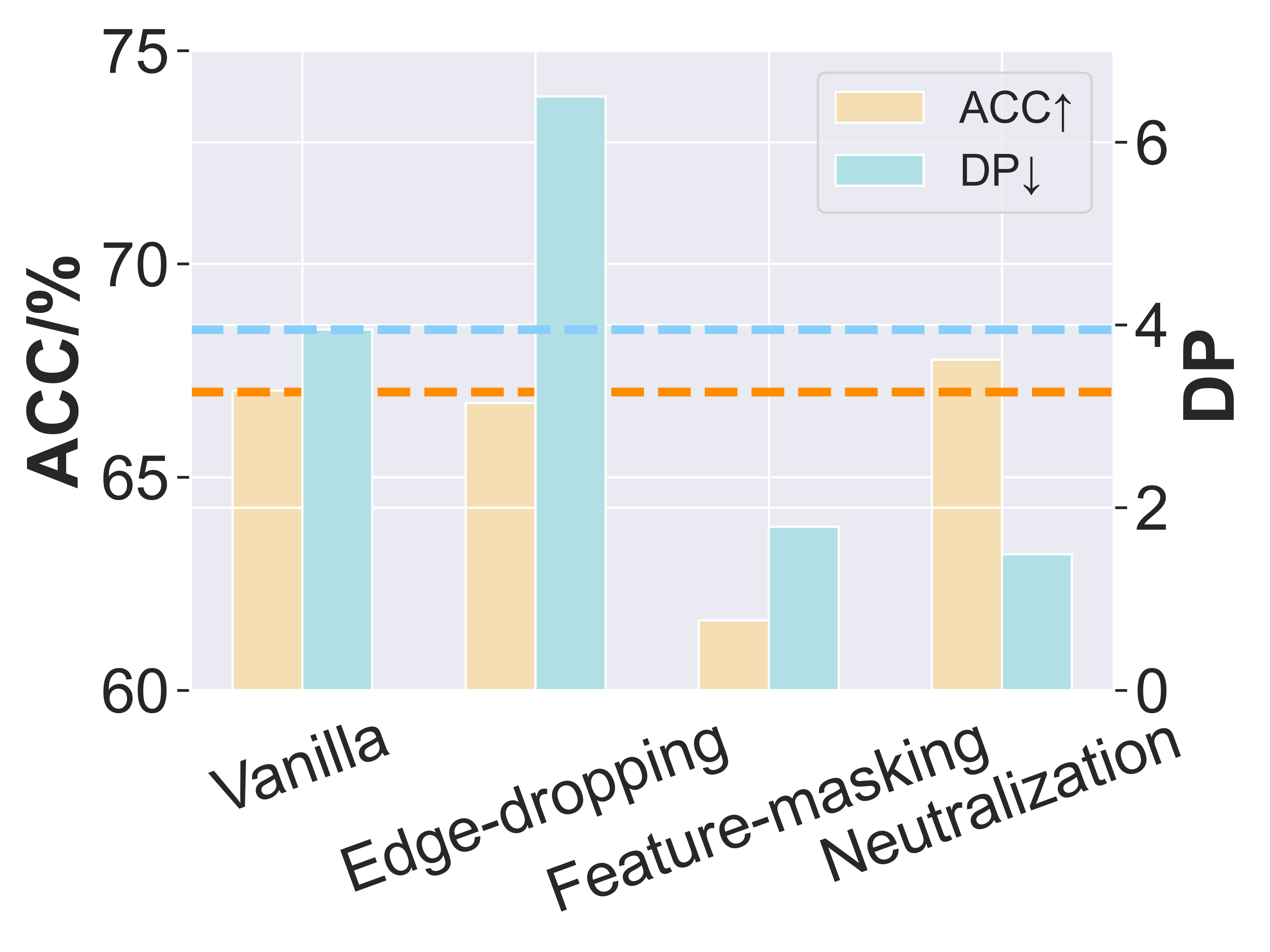}
  }
  \caption{Motivation verification on Pokec datasets. Compared with vanilla GNN without fairness consideration, filtering-based methods, either edge-dropping~\cite{nifty} or feature-masking~\cite{fairv}, always have a trade-off between accuracy (ACC$\uparrow$) and fairness (DP$\downarrow$). While our method can improve both.}
  \label{fig:cmp}
\end{figure}

Generally, the biases in GNN predictions can be attributed to both node features and graph topology: (1) The raw features of nodes could be statistically correlated to the sensitive attribute, and thus lead to sensitive information leakage in encoded representations. (2) According to the homophily effects~\cite{mcpherson2001birds,la2010randomization}, nodes with the same sensitive attribute tend to link with each other, which will make the node representations in the same sensitive group more similar during message passing. 


\begin{figure*}
    \centering
    \includegraphics[scale=0.35]{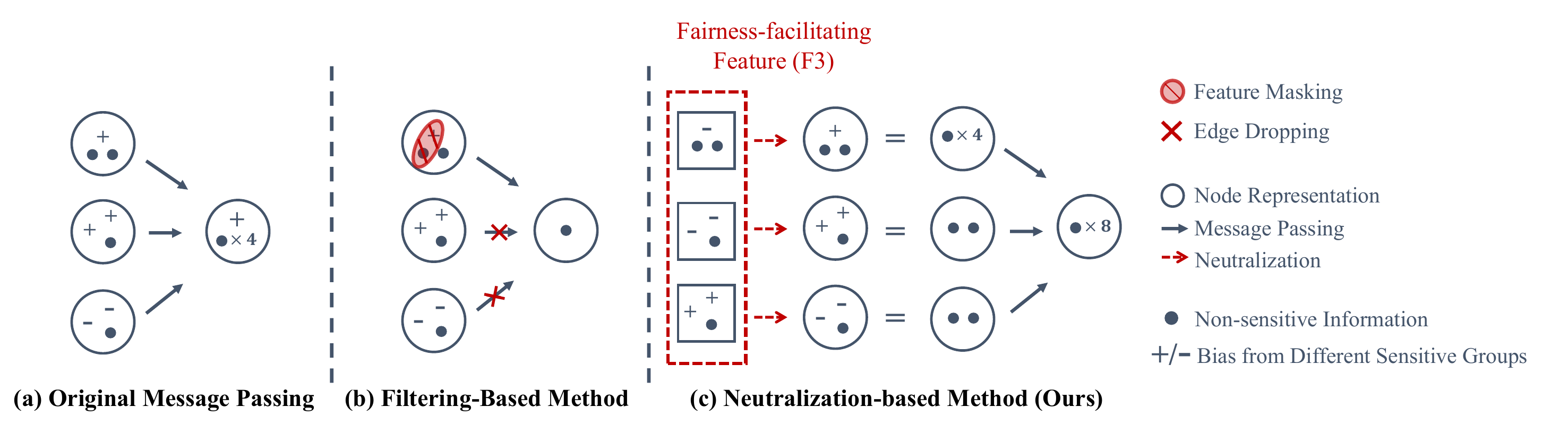}
    \caption{Motivation illustration of sensitive information neutralization. Here we assume binary sensitive groups denoted by +/-, and the numbers of +/- indicate the intensity of sensitive information leakage in node representations. (a) Message passing computation will aggregate both non-sensitive feature information (dot symbols) and sensitive biases (+/- symbols); (b) Current SOTA methods are usually \textit{filtering-based} (\textit{e.g.,} edge dropping or feature masking), which may lose much non-sensitive information; (c) Our proposed \textit{neutralization-based} strategy introduces \textit{F3} to statistically neutralize the sensitive bias and provide extra non-sensitive information.}
    \label{fig:motivation}
\end{figure*}

To address the issue of sensitive biases, researchers have introduced fairness considerations into GNNs~\cite{nifty,fairgnn,bose2019compositional,fairv,edits}. Recent state-of-the-art (SOTA) methods often attempt to mitigate the impact of sensitive information by applying heuristic or adversarial constraints to filter it out from inputs or representations. For instance, NIFTY~\cite{nifty} employs counterfactual regularizations to perturb node features and drop edges. FairVGNN~\cite{fairv}, aided by adversarial discriminators, learns adaptive representation masks to exclude sensitive-relevant information. Nevertheless, as depicted in Figure \ref{fig:cmp} and ~\ref{fig:motivation}(b), we contend that such \textit{filtering-based} strategies may also filter out some non-sensitive feature information, leading to a sub-optimal balance between accuracy and fairness. 

In light of this, we propose an alternative \textit{neutralization-based} paradigm,  as shown in Figure~\ref{fig:motivation}(c).  The core idea is to introduce extra \textit{Fairness-facilitating Features (F3)} to node features or representations so that the sensitive biases (+/- symbols) can be neutralized. The \textit{F3} are also expected to provide additional non-sensitive feature information (dot symbols), thus enabling a better trade-off between predictive performance and fairness. Specifically, we show how message passing exacerbates sensitive biases\footnote{The claim that message passing or feature propagation will intensify the sensitive biases has been mentioned~\cite{fairv,jiang2022fmp} or empirically validated \cite{edits} in previous studies. Here we have a different derivation that directly motivates our \textit{neutralization-based} design.}, and accordingly conclude that node features or representations can be debiased before message passing by emphasizing the features of each node's heterogeneous neighbors (neighbors with different sensitive attributes) as \textit{F3}. But some nodes in real-world graphs have very few or even no heterogeneous neighbors, which makes the calculation of \textit{F3} infeasible or very uncertain. Therefore, we propose to train an estimator to predict the average features or representations of a node's heterogeneous neighbors given its own feature. In this way, nodes with rich heterogeneous neighbors can transfer their knowledge to other nodes through the estimator. We name our method as \textit{FairSIN}, and further present three implementation variants from both data-centric and model-centric perspectives. Experimental results on five benchmark datasets with three different GNN backbones demonstrate the motivation and effectiveness of our proposed method.

Our contributions are as follows: (1) We present a novel \textit{neutralization-based} paradigm for learning fair GNNs, which introduces \textit{Fairness-facilitating Features (F3)} to node features/representations for debiasing sensitive attributes and providing additional non-sensitive information. (2) We show that \textit{F3} can be implemented by emphasizing the features of each node's heterogeneous neighbors, and further propose three effective variants of FairSIN. (3) Experimental results show that the proposed FairSIN can reach a better trade-off between predictive performance and fairness compared with recent SOTA methods.




\section{Related Work}
\label{2}
\paragraph{Graph Neural Networks.}
Graph-structured data widely exists in various real-world applications. To handle this type of non-Euclidean data, graph neural networks are designed for representation learning of nodes/edges/graphs, enabling a wide range of downstream tasks. For example, Graph Convolutional Network (GCN)~\cite{gcn} uses convolutional operations to perform layer-by-layer abstraction and refinement of node features. Graph Isomorphism Network (GIN)~\cite{gin} is a method proposed to have more discriminative power for graph structures and make GNN as powerful as WL-test~\cite{shervashidze2011weisfeiler}. GraphSAGE~\cite{sage} is an inductive representation learning method that can be used for large-scale data, and Graph Attention Network (GAT)~\cite{gat} is an attention-based method that assigns different weights to different neighbors during message passing. These methods have shown outstanding performance in various graph-based applications.
    
\paragraph{Fairness in Graph Neural Networks.} 
Fairness issues in machine learning models have gained increasing attention from both academia \cite{dong2023fairness,chouldechova2018frontiers,sun2019mitigating,mehrabi2021survey,field2021survey} and industry \cite{holstein2019improving}. In terms of GNNs, there are different fairness definitions proposed in the literature \cite{dp,eo,individual,cf,degree,wang2022uncovering}. Among them, group fairness is one of the most popular notion \cite{dp,eo}, which aims at providing equal predictions for all demographic groups without any biases or discrimination. Due to the presence of sensitive-relevant features, GNNs may inadvertently perpetuate or amplify biases and discrimination against certain sensitive groups. 

Improving group fairness in GNNs has attracted much attention over the last five years~\cite{li2021dyadic,bose2019compositional,dong2023fairness,fairwalk,laclau2021all,fisher2020debiasing}. Recent SOTA methods usually employ feature masking or topology modification to filter out sensitive biases during message passing. For feature masking, \cite{bose2019compositional} leverages adversarial learning to enforce compositional fairness constraints on graph embeddings for multiple sensitive attributes filtering. FairVGNN~\cite{fairv} uses a mask generator to filter out channels with high correlation to sensitive attributes. While others have investigated how to drop edges with high bias \cite{dong2022structural,fairdrop}. For example, REFEREE \cite{dong2022structural} provides structural explanations of topology bias on how to improve fairness. FairDrop~\cite{fairdrop} adopts edge masking to counter-act homophily. It is worth noting that some methods \cite{edits,ling2023learning} not only drop edges but also add new ones. However, considering the addition of new edges has to compute the similarity between all node pairs that are not connected, thus introducing a time and space complexity of $\mathcal{O}(n^2)$. This could be computationally expensive and thus edge dropping is a more practical choice. Also, some methods consider both feature masking and topology modification~\cite{edits,ling2023learning,fairaug}. Such \textit{filtering-based} methods based on feature masking or edge dropping will unavoidably lead to the loss of useful non-sensitive information. Therefore, we propose a novel \textit{neutralization-based} method that introduces \textit{F3} to statistically neutralize the sensitive bias and provide extra non-sensitive information. 

\section{Methodology}
In this section, we will elaborate on the details of our proposed FairSIN, which employs the features/representations of heterogeneous neighbors to simultaneously neutralize sensitive biases and incorporate extra non-sensitive information. 

\subsection{Preliminaries}

\paragraph{Notations.} Let $\mathcal{G} = (\mathcal{V},\mathcal{E})$ be a graph with node set $\mathcal{V}$ and edge set $\mathcal{E}$. 
The adjacency matrix $\mathbf{A} \in \mathbb{R}^{|\mathcal{V}| \times |\mathcal{V}|}$ represents the connectivity between nodes, where $\mathbf{A}_{ij}=1$ if there is a directed edge between nodes $v_i$ and $v_j$, and $\mathbf{A}_{ij}=0$ otherwise. The node feature matrix $\mathbf{X} \in \mathbb{R}^{|\mathcal{V}| \times d}$ contains the feature vectors for each node, where $\mathbf{x}_{i} \in \mathbb{R}^d$ is the feature vector for node $v_i\in \mathcal{V}$. Besides, each node $v_i$ also has a categorical sensitive attribute $s_{i}\in \mathcal{S}$. 

\paragraph{Task Definition.} In this paper, we consider the benchmark task as previous work~\cite{fairv,nifty,fairgnn} did, \textit{i.e.,} the semi-supervised node classification task. Formally, given graph $\mathcal{G}$, node features $\mathbf{X}$ and labeled node set $\mathcal{V}^L \subset \mathcal{V}$, we need to build a model to predict the label $\hat{y}\in \mathcal{Y}$ for every node in the unlabeled node set $\mathcal{V}^U=\mathcal{V}\setminus \mathcal{V}^L$. A typical design of the model is to combine a GNN encoder and a classifier. The classification performance can be measured by aligning the ground truth labels $Y$ and predicted ones $\hat{Y}$. In terms of fairness, the goal is to weaken the dependency level between predicted labels $\hat{Y}$ and sensitive attributes $S$, without losing much classification accuracy. In practice, many fair representation learning methods ~\cite{edwards2015censoring,fairgnn,bose2019compositional,liao2019learning} will minimize the dependency between node representations and sensitive attributes instead.

\subsection{Theoretical Analysis}
\label{sec:theory}
In this subsection, we will introduce our motivation of neutralizing sensitive information from a theoretical perspective. Firstly, we treat node features and graph topology as random variables, and describe a generative process to model the dependency among them. Similar generative process has been done in \cite{wang2022unbiased}. Then we propose to measure the sensitive information leakage by the conditional entropy between sensitive attributes and node representations. Finally, we show how the message passing computation exacerbates the leakage problem of sensitive information. Note that there is a slight abuse of notations about random variables and samples in this subsection.

\paragraph{Generative Process.} Here we describe a graph generation process with the following two steps: (1) For each node $v_i$, we draw its features and sensitive attribute from a joint prior distribution $(x_i,s_i)\sim \textit{prior}$. For simplicity, we assume that the sensitive attribute is binary as previous work did~\cite{fairv,nifty,fairgnn}, and use $\bar{s}$ to denote the opposite counterpart of $s$. (2) To obtain graph $\mathcal{G}$, each node $v_i$ samples its in-degree neighbor set $\mathcal{N}_i$ by the \textit{homophily} assumption\cite{mcpherson2001birds, wang2022unbiased}, where nodes with similar features or sensitive attributes are more likely to get connected. We name the neighbors with same/different sensitive attributes as homogeneous/heterogeneous neighbors, respectively.

We use $P_i^\textit{same}$/$P_i^\textit{diff}$ to represent the probability that $v_i$ samples a homogeneous/heterogeneous neighbor. According to the homophily assumption, $P_i^\textit{same}>P_i^\textit{diff}$. Then we denote the average feature of $v_i$'s in-degree neighbors as $x_{i}^\textit{neigh}$. The average features of $v_i$'s homogeneous/heterogeneous in-degree neighbors are written as $x_{i}^\textit{same}$/$x_{i}^\textit{diff}$. Thus $x_{i}^\textit{neigh}=P_i^\textit{same}x_{i}^\textit{same}+P_i^\textit{diff}x_{i}^\textit{diff}$.


\paragraph{Quantifying Sensitive Information Leakage.} In this work, we focus on alleviating sensitive biases in node representations, and use the conditional entropy between sensitive attributes and node representations as the measurement. Without loss of generality, we take raw features $x$ as node representations, and compute the conditional entropy as
\begin{equation}
\mathcal{H}(s|x)=-\mathbb{E}_{(x,s)\sim prior} \log P(s|x),
\end{equation}
where $P(s|x)$ is a predictor that estimates sensitive attributes given node representations. When node representations have more sensitive information leakage, the predictor will be more accurate and the entropy will get smaller. 

In practical applications, it becomes necessary to approximate the ground truth predictor $P(s|x)$. Specifically, we adopt a linear intensity function $\mathcal{D}_\theta$ with the parameter $\theta$ to define the predictive capability, satisfying the conditions:
$\mathcal{D}_\theta(s|x) \sim \mathcal{N}(\mu_{c}, \sigma^2)$ and $\mathcal{D}_\theta(\bar{s}|x) \sim \mathcal{N}(\mu_{ic}, \sigma^2)$,
where $(x,s)\sim \textit{prior}$ and $\mathcal{N}(\cdot,\cdot)$ is the Gaussian distribution. $\mu_c > \mu_{ic}$ indicates that $\mathcal{D}_\theta$ is more likely to assign larger intensity score to the true sensitive attribute given node representations. The larger $\mu_c-\mu_{ic}$ is, the stronger the inference capability of $\mathcal{D}_\theta$. In contrast, $\mu_c =\mu_{ic}$ means that $\mathcal{D}_\theta$ can not distinguish the sensitive attribute from representations.

Then we can define the parameterized predictor $\hat{P}_\theta(s|x)$ by normalizing the intensity function $\mathcal{D}_\theta$. 

\paragraph{Message Passing Can Exacerbate Sensitive Biases.} Now we will straightforwardly provide an exposition from the perspective of sensitive information neutralization that the message passing computation may lead to more serious sensitive information leakage problem.

\begin{theorem}
\label{theo1}
Assume that node representations are biased and can be identified by the predictor, \textit{i.e.,} $\mu_c > \mu_{ic}$. For node $v_i$, we consider a message passing process that updates $x_i$ by $x_i'=x_i+x_{i}^\textit{neigh}$. Then we have 
\begin{equation}
    \mathbb{E} \{\mathcal{D}_\theta(s_i|x_i')-\mathcal{D}_\theta(\bar{s_i}|x_i')\}>\mathbb{E} \{\mathcal{D}_\theta(s_i|x_i)-\mathcal{D}_\theta(\bar{s_i}|x_i)\},
\end{equation}
which means that the predictor $\hat{P}_\theta$ can identify the sensitive attributes more accurately. 
\end{theorem}
The details and the proof of Theorem 1 can be found in the Appendix.

\paragraph{Summary.}Therefore, to alleviate the sensitive biases, we can either modify the graph structure to decrease $P_i^\textit{same}-P_i^\textit{diff}$ or modify the node features before message passing to decrease $\mu_c - \mu_{ic}$. Both solutions actually emphasize the features of each node's heterogeneous neighbors, and can be seen as introducing extra \textit{F3} into node representations for sensitive information neutralization.


\subsection{Implementations of FairSIN}
Inspired by the above motivation, we will present three implementation variants from both data-centric and model-centric perspectives. We name our method as \textit{Fair GNNs via Sensitive Information Neutralization} (FairSIN).

\paragraph{Data-centric Variants.} For data-centric implementation, we will employ a pre-processing manner, and modify the graph structure or node features before the training of GNN encoder. 

(1) In terms of graph modification, we can simply change the edge weights in the adjacency matrix: 
\begin{equation}
\mathbf{A}_{ij}=
\begin{cases}
\quad   1+\delta, \quad \text{if $(v_i,v_j)\in \mathcal{E}$ and $s_i\neq s_j$}       \\
\quad   1, \quad\quad\;\; \text{if $(v_i,v_j)\in \mathcal{E}$ and $s_i= s_j$}       \\
\quad 0, \quad\quad\;\; \text{if $(v_i,v_j)\not\in \mathcal{E}$}       \\
\end{cases},
\end{equation}
where $\delta>0$ is a hyper-parameter. We name this variant as FairSIN-G.

(2) In terms of feature modification, we first compute the average feature of each node $v_i$'s heterogeneous neighbors as $\mathbf{x}_i^\textit{diff}=\frac{1}{|\mathcal{N}_i^\textit{diff}|}\sum_{v_j\in \mathcal{N}_i^\textit{diff}} \mathbf{x}_j$, where $\mathcal{N}_i^\textit{diff}$ is the heterogeneous neighbor set of $v_i$. Here $\mathbf{x}_i^\textit{diff}$ can also be seen as the expectation estimation of the random variable ${x}_i^\textit{diff}$ defined in previous subsection.

However, some nodes in real-world graphs have very few or even no heterogeneous neighbors, which makes the calculation of $\mathbf{x}_i^\textit{diff}$ infeasible or very uncertain. To address this issue, we propose to train a multi-layer perceptron (MLP)\footnote{From an empirical standpoint, MLPs are simple and already qualified to achieve our desired outcomes. We will explore more sophisticated architectures in future work.} to estimate $\mathbf{x}_i^\textit{diff}$:
\begin{equation}
    \mathcal{L}_{F} = \frac{1}{|\mathcal{V}|}\sum_{i:|\mathcal{N}_i^\textit{diff}|\geq 1}\Vert\operatorname{MLP}_\phi(\mathbf{x}_i) - \mathbf{x}_i^\textit{diff}\Vert^2.
\end{equation}
By minimizing the above Mean Squared Error (MSE) loss, nodes with rich heterogeneous neighbors can transfer their knowledge to other nodes through the MLP. Then we neutralize each node $v_i$'s feature as $\Tilde{\mathbf{x}}_i=\mathbf{x}_i+\delta \operatorname{MLP}_\phi(\mathbf{x}_i)$, and name this variant as FairSIN-F.

\paragraph{Model-centric Variants.} The model-centric variant further extends FairSIN-F by jointly learning the MLP$_\phi$ and GNN encoder. Given a $K$-layer GNN, we denote the representation matrix of all nodes in the $k$-th layer as $\mathbf{H}^k$. Similar to FairSIN-F, we can conduct the neutralization operation at every layer: 
\begin{equation}
\begin{aligned}
\Tilde{\mathbf{H}}^k &= \mathbf{H}^k + \delta^k \operatorname{MLP}_\phi^k(\mathbf{H}^k), \\
\mathbf{H}^{k+1} &= \operatorname{MessagePassing}(\Tilde{\mathbf{H}}^k),
\end{aligned}
\end{equation}
where $\mathbf{H}^0=\mathbf{X}$ is the node feature matrix, $\delta^k$ and $\operatorname{MLP}_\phi^k$ can be customized for each layer. We denote the MSE loss in each layer $k$ as $\mathcal{L}_{F}^k$.

Following recent SOTA methods on fair GNNs, we also introduce a discriminator module to impose extra fairness constraints on the encoded representations. Specifically, we use another MLP$_\psi$ to implement the discriminator, and let it predict the sensitive attribute based on the final representation encoded by GNN. We use binary cross-entropy (BCE) loss $\mathcal{L}_D$ to train the discriminator, and ask the GNN encoder and MLP$_\phi$ to maximize $\mathcal{L}_D$ as adversaries. Besides, we denote the cross-entropy loss of downstream classification task as $\mathcal{L}_T$. For parameter training, we iteratively perform the following steps: (1) update each MLP$_\phi^k$ by minimizing $\mathcal{L}_F^k-\mathcal{L}_D$; (2) update GNN encoder by minimizing $\mathcal{L}_T-\mathcal{L}_D$; and (3) update discriminator MLP$_\psi$ by minimizing $\mathcal{L}_D$. We consider this variant as our full model FairSIN. 


\subsection{Discussion}
\paragraph{How Does FairSIN Work?} Recall that the neutralized feature $\Tilde{\mathbf{x}}_i$ is the expectation estimation of the random variable $x_i+\delta{x}_i^\textit{diff}$. Similar to the theorem, we have
\begin{equation}
\begin{aligned}
    &\mathbb{E} \{\mathcal{D}_\theta(s_i|x_i+\delta{x}_i^\textit{diff})-\mathcal{D}_\theta(\bar{s_i}|x_i+\delta{x}_i^\textit{diff})\} \\ 
    &=(1 - \delta)(\mu_c -\mu_{ic})<\mathbb{E} \{\mathcal{D}_\theta(s_i|x_i)-\mathcal{D}_\theta(\bar{s_i}|x_i)\},
\end{aligned}
\end{equation}
where $\delta\in (0,1]$. Thus it is harder to infer the sensitive attribute from the neutralized feature than the raw feature. In practice, we relax the range of $\delta$ to a wider range for more flexible tuning.

Here we present an empirical verification of our theory. We consider four groups of node features, including the raw feature ${\mathbf{x}}_i$, raw feature + message passing ${\mathbf{x}}_i+\frac{1}{|\mathcal{N}_i|}\sum_{v_j\in \mathcal{N}_i} \mathbf{x}_j$, neutralized feature $\Tilde{\mathbf{x}}_i$, neutralized feature + message passing $\Tilde{\mathbf{x}}_i+\frac{1}{|\mathcal{N}_i|}\sum_{v_j\in \mathcal{N}_i} \Tilde{\mathbf{x}}_j$. For each group of features, we train a sensitive attribute predictor as in Section~\ref{sec:theory}, and use average $\hat{P}_\theta(s|x)$ as the measurement score\footnote{We did not use the conditional entropy since the log operation has numerical instability issues in practice.}. A larger score indicates more serious sensitive biases of node representations.

\begin{figure}[t]
    \centering
    \includegraphics[scale=0.33]{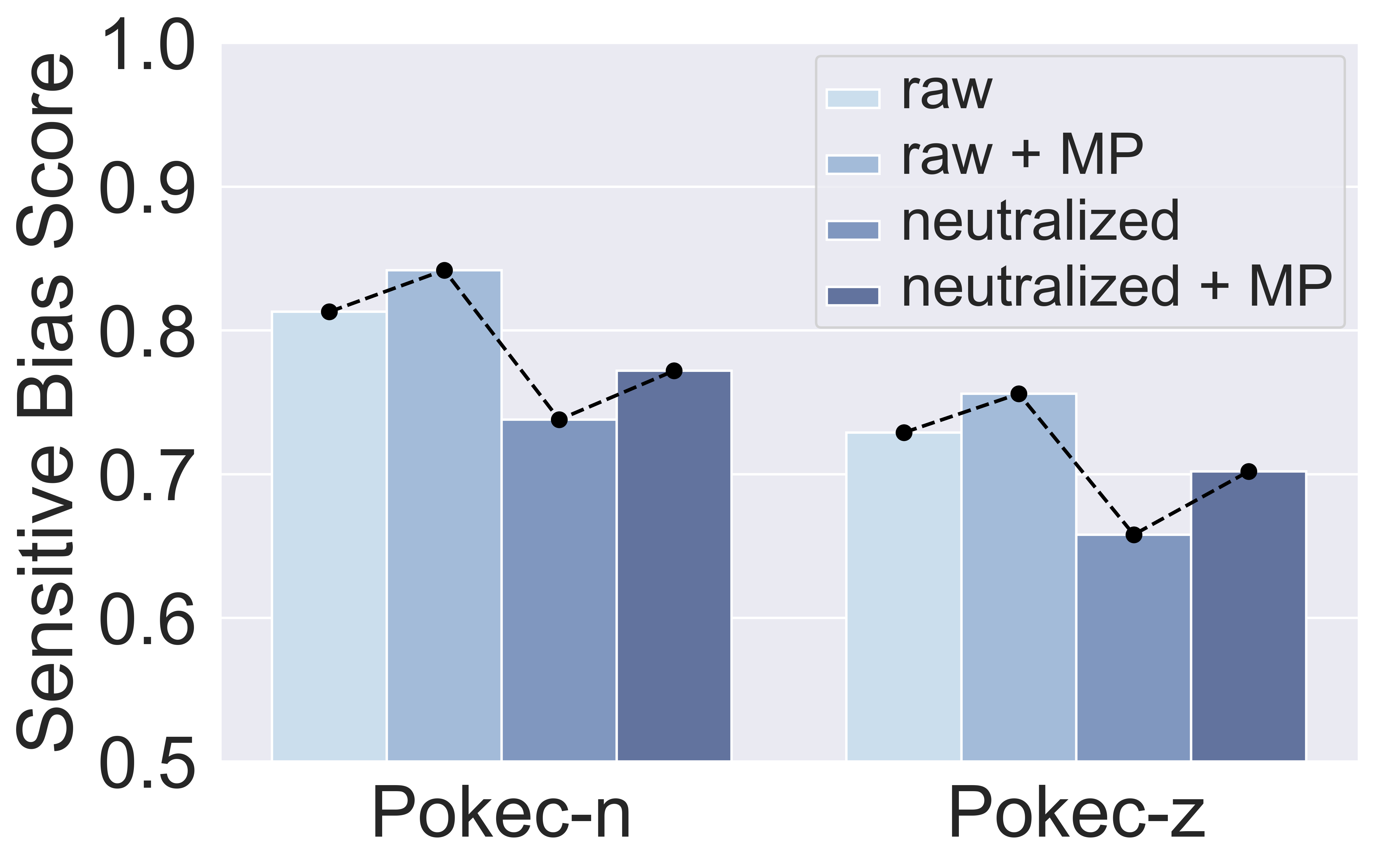}
    \caption{Sensitive biases in four groups of features. The biases are measured by average $\hat{P}_\theta(s|x)$, and larger scores indicate more serious sensitive leakage in the representations.}
    \label{fig:score}
\end{figure}

From Figure~\ref{fig:score}, we can see that message passing enlarges the sensitive biases for both raw and neutralized features, which can validate our theoretical analysis. Also, neutralized features have much less sensitive information leakage, demonstrating the effectiveness of our \textit{F3}. 

Moreover, the estimated feature of heterogeneous neighbors $\operatorname{MLP}_\phi(\mathbf{x}_i)$ can provide additional information when calculating representations, especially for the nodes with few heterogeneous neighbors. Therefore, our method can reach a better performance-fairness trade-off than previous fair GNN methods.



\paragraph{Data-centric v.s. Model-centric Variants.} As pre-processing methods, data-centric variants are task-irrelevant and thus can be employed for various downstream scenarios. For example, we can debias a graph dataset by neutralizing node features in advance, and then graph machine learning algorithms can be trained as usual. Data-centric variants are also more computationally efficient. The model-centric variant is also model-agnostic, and can be combined with any GNN encoders. It allows to further neutralize the internal representations in each GNN layer, and enables additional fairness constraint from an adversarial discriminator. Different parts of the model can learn and improve together, thereby achieving better accuracy and fairness.




\begin{table}[t]
\renewcommand{\arraystretch}{1} 
\small 
\begin{tabular}{l|cccl}
\specialrule{2pt}{0pt}{0pt}
Dataset                      & Bail        &Pokec-n         & Pokec-z          \\ \specialrule{1.5pt}{0pt}{0pt}
\# Nodes                      & 18,876            & 66,569         & 67,797         \\
\# Features                   & 18            & 266             & 277              \\
\# Edges                      & 321,308        & 729,129      & 882,765       \\
Node label                        & Bail decision & Working field & Working field   \\
Sens. Attri.                        & Race        & Region            & Region            \\

Avg. Deg.                  & 34.04        & 16.53          & 19.23          \\
Avg. H-Deg.          & 15.79          & 0.73           & 0.90         \\
 \specialrule{2pt}{0pt}{0pt}
\end{tabular}
\caption{Dataset statistics. ``H-'' means ``heterogeneous''.}
\label{tab:dataset}
\end{table}

\begin{table*}[h]
	\centering
        \renewcommand{\arraystretch}{1.2} 
\footnotesize        
\begin{tabular}{@{}c@{}|c|c@{\hspace{0.1cm}}c@{\hspace{0.1cm}}c|c@{\hspace{0.1cm}}c@{\hspace{0.1cm}}c|c@{\hspace{0.1cm}}c@{\hspace{0.1cm}}c}
\specialrule{1.5pt}{0pt}{0pt}
\multirow{2}{*}{Encoder} & \multirow{2}{*}{Method} & \multicolumn{3}{c|}{Bail}                                                         & \multicolumn{3}{c|}{Pokec\_n}                                                         & \multicolumn{3}{c}{Pokec\_z}                                                            \\ \cline{3-11}
                         &                                         & ACC↑                & DP↓                & EO↓                                & ACC↑                & DP↓                & EO↓                                 & ACC↑                & DP↓                & EO↓               \\ \hline \hline
\multirow{10}{*}{GCN}      & vanilla                                                  & 87.55±0.54               & 6.85±0.47               & 5.26±0.78                            & 68.55±0.51               & 3.75±0.94               & 2.93±1.15                            & 66.78±1.09         & 3.95±1.03               & 2.76±0.95               \\
                         & FairGNN                                                 & 82.94±1.67               & 6.90±0.17               & 4.65±0.14                            & 67.36±2.06               & 3.29±2.95               & 2.46±2.64                             & \underline{67.65±1.65}               & 1.87±1.95               & 1.32±1.42               \\
                         & EDITS                                                 & 84.49±2.27               & 6.64±0.39               & 7.51±1.20                                   & OOM                      & OOM                     & OOM                                         & OOM                      & OOM                     & OOM                     \\
                         & NIFTY                                                     & 82.36±3.91               & 5.78±1.29               & 4.72±1.08                             & 67.24±0.49               & 1.22±0.94               & 2.79±1.24                          & 66.74±0.93               & 6.50±2.16               & 7.64±1.77               \\ 
                         & FairVGNN                                                 & 84.73±0.46               & 6.53±0.67               & 4.95±1.22                             & 66.10±1.45               & 1.69±0.79               & 1.78±0.70                             & 61.64±4.72               &  \underline{1.79±1.22}         & \underline{1.25±1.01}         \\ \cline{2-11}
                         & FairSIN-G                                              & 85.57±1.08               & 6.57±0.29               & 5.55±0.84                     & 68.22±0.39               & 2.56±0.60               & 1.69±1.29                            & 65.73±1.76               & 3.53±1.20               & 2.42±1.43               \\
                         & FairSIN-F                                            & \underline{87.61±0.83}         & \underline{5.54±0.40}               & \underline{3.47±1.03}                       & 67.96±1.54               & \underline{1.16±0.90}         & \underline{0.98±0.70}              & 66.38±1.39               & 2.53±0.97               & 2.03±1.23               \\
                         & FairSIN w/o N.                              & 87.26±0.17               & 5.93±0.04               & 4.30±0.20                            & 68.35±0.62               & 2.51±1.99               & 2.36±1.89                             & 65.87±1.34               & 1.98±1.01               & 1.87±0.64               \\
                         & FairSIN w/o D.                                  & 87.40±0.15               & 5.65±0.40         & 4.63±0.52                    & \underline{68.74±0.33}         & 2.22±1.47               & 1.67±1.70                       & 66.42±1.52               & 2.73±1.08               & 2.37±0.69               \\
                         & \textbf{FairSIN}                                            & \textbf{87.67±0.26}      & \textbf{4.56±0.75}      & \textbf{2.79±0.89}                   & \textbf{69.34±0.32}      & \textbf{0.57±0.19}      & \textbf{0.43±0.41}                    & \textbf{67.76±0.71}      & \textbf{1.49±0.74}      & \textbf{0.59±0.50}      \\ \hline \hline
\multirow{10}{*}{GIN}      & vanilla                                                  & 83.52±0.87               & 7.55±0.51               & 6.17±0.69                         & 69.25±1.75               & 3.71±1.20               & 2.55±1.52                       & 65.83±1.31               & 1.97±1.12               & 2.17±0.48               \\
                         & FairGNN                                               & 77.90±2.21               & 6.33±1.49               & 4.74±1.64                           & 67.10±3.25               & 3.82±2.44               & 3.62±2.78                     & \underline{66.49±1.54}         & 3.53±3.90               & 3.17±3.52               \\
                         & EDITS                                                     & 73.74±5.12               & 6.71±2.35               & 5.98±3.66                                                          & OOM                     & OOM                     & OOM                     & OOM                      & OOM                     & OOM                     \\
                         & NIFTY                                                   & 74.46±9.98               & 5.57±1.11               & \textbf{3.41±1.43}                    & 66.37±1.51               & 3.84±1.05               & 3.24±1.60                            & 65.57±1.34               & 2.70±1.28               & 3.23±1.92               \\
                         & FairVGNN                                                & 83.86±1.57               & 5.67±0.76               & 5.77±0.76                             & 68.37±0.97               & 1.88±0.99               & 1.24±1.06                             & 65.46±1.22               & 1.45±1.13               & 1.21±1.06               \\ \cline{2-11}
                         & FairSIN-G                                                & 86.10±1.39               & 6.93±0.16               & 6.75±0.66                            & 67.73±1.67               & 1.98±1.54               & 1.50±1.15                           & 65.09±2.69               & 1.55±1.23               & 1.74±0.80               \\
                         & FairSIN-F                     & \underline{86.48±0.75}         & 5.95±1.85               & 5.97±2.07                      & 68.92±1.08               & \underline{1.51±1.11 }              & \textbf{0.82±0.79}                            & 65.97±0.82               & 1.45±1.15               & 1.14±0.73               \\
                         & FairSIN w/o N.                             & 85.27±0.70               & 7.21±0.39               & 6.75±0.55                            & 68.92±1.13               & 2.81±1.91               & 2.12±1.30                            & 65.04±1.56               & 2.19±1.96               & 1.23±0.92               \\
                         & FairSIN w/o D.                             & 86.44±0.80               & \underline{4.38±1.48}               & 4.23±1.88                             & \textbf{70.04±0.80}      & 2.44±1.50                & 1.63±1.24                             & 65.58±0.71               & \underline{1.40±0.67}         & \underline{1.12±0.24}         \\
                         & \textbf{FairSIN}                                         & \textbf{86.52±0.48}      & \textbf{4.35±0.71}      & \underline{4.17±0.96}             & \underline{69.58±0.57}         & \textbf{1.11±0.31}      & \underline{0.97±0.59}                   & \textbf{66.74±1.56}      & \textbf{0.64±0.47}      & \textbf{1.01±0.64}      \\ \hline \hline
\multirow{10}{*}{SAGE}    & vanilla                                                   & 88.13±1.12               & 1.13±0.48               & 2.61±1.16                             & 69.03±0.77               & 3.09±1.29               & 2.21±1.60                             & 66.55±0.69               & 4.71±1.05               & 2.72±0.85               \\
                         & FairGNN                                               & 87.68±0.73               & 1.94±0.82               & 1.72±0.70                        & 67.03±2.61               & 2.97±1.28               & 2.06±3.02                            & \underline{67.68±1.49}         & \underline{2.86±1.39}         & 2.30±1.33                \\
                         & EDITS                                                    & 84.42±2.87               & 3.74±3.54               & 4.46±3.50                                   & OOM                     & OOM                     & OOM                     & OOM                      & OOM                     & OOM                     \\
                         & NIFTY                                                    & 84.11±5.49               & 5.74±0.38               & 4.07±1.28                            & 68.48±1.11               & 3.84±1.05               & 3.90±2.18                             & 66.68±1.45               & 6.75±1.84               & 8.15±0.97               \\
                         & FairVGNN                                                 & 88.41±1.29               & 1.14±0.67               & 1.69±1.13                           & 68.50±0.71               & \underline{1.12±0.98}         & 1.13±1.02                             & 66.39±1.95               & 4.15±1.30               & 2.31±1.57               \\ \cline{2-11}
                         & FairSIN-G                                           & \textbf{88.79±1.08}      & 3.97±0.92               & \underline{1.70±0.66}                      & 69.11±0.62               & 2.00±1.13               & 1.66±0.70                  & 66.19±1.49               & 4.96±0.25               & 2.90±1.21               \\
                         & FairSIN-F                                                   & 88.51±0.16               & 0.67±0.33              & 1.85±0.50                          & \underline{69.28±0.98}         & 1.80±0.46               & 1.62±0.84                            & 66.99±1.06               & 3.25±1.00               & 1.89±0.79               \\
                         & FairSIN w/o N.                                & 87.70±0.28               & \underline{0.64±0.40}               & 2.21±0.22                            & 68.77±0.62               & 2.35±0.99               & 1.71±0.99                              & 67.39±1.05               & 2.92±1.69               & 1.79±1.16               \\
                         & FairSIN w/o D.                               & 88.46±0.19               & 0.82±0.51               & 2.12±0.55                             & \textbf{69.65±0.32}      & 1.91±0.82               & \underline{1.09±1.12}                     & 66.78±0.83               & 3.92±1.02               & \underline{1.62±0.68}         \\
                         & \textbf{FairSIN}                                            & \underline{88.74±0.42}         & \textbf{0.58±0.60}      & \textbf{1.49±0.34}          & 69.12±1.16               & \textbf{1.04±0.83}      & \textbf{1.04±0.42}             & \textbf{67.95±0.79}      & \textbf{1.74±0.73}      & \textbf{0.68±0.65}      \\ \specialrule{1.5pt}{0pt}{0pt}
\end{tabular}
 \caption{Comparison among SOTA methods and different variants of FairSIN.  ({Bold}: the best; {underline}: the runner-up.)}
 \label{tab:main}
\end{table*}

\section{Experiments}
\label{Experiments}
In this section, we thoroughly evaluate and analyze the effectiveness of our proposed method. Specifically, we follow the experimental setup of \cite{fairv, edits}, and consider both fairness and accuracy metrics across multiple datasets. 
Our experiments are conducted to answer the following research questions (RQs):

\textbf{RQ1}: How effective is our proposed method compared with SOTA graph fairness methods? 
\textbf{RQ2}: How does each module of our proposed method contribute to the final performance? 
\textbf{RQ3}: How does the hyper-parameter $\delta$ influence the performance? 
\textbf{RQ4}: How does the time cost of our method compared with other baselines? 

\subsection{Experimental Settings}

\paragraph{Datasets.}
Following the approaches proposed in \cite{fairv,edits}, we evaluate FairSIN and baseline methods on five real-world benchmark datasets\footnote{Due to space limitation, the statistics and results of German and Credit datasets are in the Appendix.}: German \cite{asuncion2007uci}, Credit \cite{yeh2009comparisons}, Bail \cite{jordan2015effect} and Pokec-n/Pokec-z \cite{pokec}. These datasets have been extensively used in previous studies on graph fairness learning, and cover a diverse range of domains, including finance, criminal justice and social network. We provide the dataset statistics in Table~\ref{tab:dataset}. 

\paragraph{GNN Backbones.}
In our experiments, we employ three commonly used graph neural networks (GNNs) as the backbone of our encoder: GCN \cite{gcn}, GIN \cite{gin}, and GraphSAGE \cite{sage}. These encoders have been widely adopted by the research community and have demonstrated strong performance on various graph-related tasks.

\paragraph{Baselines.}
We compare our methods with the following state-of-the-art fair node representation learning methods. FairGNN~\cite{fairgnn}: a debiasing method based on adversarial training. EDITS~\cite{edits}: an augmentation-based method minimizing discrimination between different sensitive groups by pruning the graph topology and node features. NIFTY~\cite{nifty}: a method that integrates feature perturbation and edge dropping to enforce counterfactual fairness constraints by maximizing the similarity between augmented and counterfactual graphs. FairVGNN~\cite{fairv}: a framework preventing sensitive attribute leakage by masking sensitive-correlated channels and adaptively clamping weights. All baselines are implemented based on the given three GNN backbones. 

\paragraph{Evaluation Metrics.}  
To assess the performance of downstream classification task, we employ F1 score and accuracy as the metrics. To evaluate group fairness, we adopt \textit{Demographic (Statistical) Parity (DP)}~\cite{dp} and \textit{Equal Opportunity (EO)} ~\cite{eo} as previous studies~\cite{nifty,fairgnn,fairv,edits}. Note that a model with lower DP and EO implies better fairness. 
    
\paragraph{Implementation Details.}
For our proposed method, we leverage a 3-layer MLP to predict the features or representations of heterogeneous neighbors. Specifically, we adopt an Adam optimizer for MLP with weight decay in \{0.001, 0.0001, 0\} and tune the learning rate in \{0.1,0.01,0.001\}. The dropout rate is in \{0.2, 0.5, 0.8\}. In addition, we tune the coefficient hyper-parameter $\delta$ in our proposed method over the range of [0,10]. For GNN encoders, we use the same settings as \cite{fairv}. We report mean and standard deviation over five runs with different random seeds. All our experiments are run on a single GPU device of GeForce GTX 3090 with 22 GB memory.

\subsection{Main Results (RQ1)}
\paragraph{Effectiveness of Model-centric Variant FairSIN.} Here we present the results of FairSIN to demonstrate that our neutralization-based strategy can achieve a better trade-off than SOTA methods. As shown in Table~\ref{tab:main}, FairSIN has both the best overall classification performance and group fairness under different GNN encoders. In terms of fairness, FairSIN respectively reduces DP and EO by 63.29\% and 33.82\%, compared with the best performed baseline. Additionally, since the \textit{F3} can introduce extra neighborhood information for each node, in many cases FairSIN can even outperform the vanilla encoder in accuracy metrics. Compared to Bail, Pokec-n/Pokec-z have very few heterogeneous neighbors. Hence, the improvement achieved by FairSIN is more pronounced on the Pokec dataset, aligning with our motivation and model design.


\paragraph{Effectiveness of Data-centric Variants FairSIN-G and FairSIN-F.}
Here we can compare our proposed data-centric variants with a previous pre-processing method EDITS~\cite{edits} as well as the vanilla encoder. From Table~\ref{tab:main}, we can see that both FairSIN-G and FairSIN-F maintain the accuracy and improve the fairness on average, which demonstrates our idea of sensitive information neutralization. Also, FairSIN-G only amplifies the weights of existing heterogeneous neighbors, which limits its capacity to furnish as extensive information as FairSIN-F. Consequently, in comparison, the predictive performance of FairSIN-G falls short when contrasted with FairSIN-F. 
It is worth noting that as a pre-processing method, FairSIN-F is only slightly worse than the model-centric variant FairSIN, and outperforms previous SOTA methods. Therefore, FairSIN-F offers a cost-effective, model-agnostic and task-irrelevant solution for fair node representation learning.


\subsection{Ablation Study (RQ2)}
To fully evaluate the effects of each component used in our proposed FairSIN, we consider two ablated models: \textit{FairSIN w/o D.} denotes the version of FairSIN without the discriminator, and \textit{FairSIN w/o N.} denotes the version of FairSIN where $\delta=0$. Experimental results are listed in Table~\ref{tab:main}. 
The relative improvement brought by \textit{F3} on Bail is not as significant as that on Pokec, since nodes in Bail dataset have almost equal number of homogeneous and heterogeneous neighbors. Broadly speaking, both \textit{F3} and the discriminator yield beneficial outcomes. However, when the discriminator is employed in isolation rather than as a constraint to guide the learning of \textit{F3}, it often leads to a decrease in predictive precision. It is worth noting that the neutralization of \textit{F3} alone, \textit{i.e.,} \textit{FairSIN w/o D.}, can already achieve a favorable trade-off between fairness and accuracy metrics, and is the most important design in our model. 

\begin{figure}[h]
  \centering
      \subfloat[{Pokec-n}]{
    \includegraphics[scale=0.15]{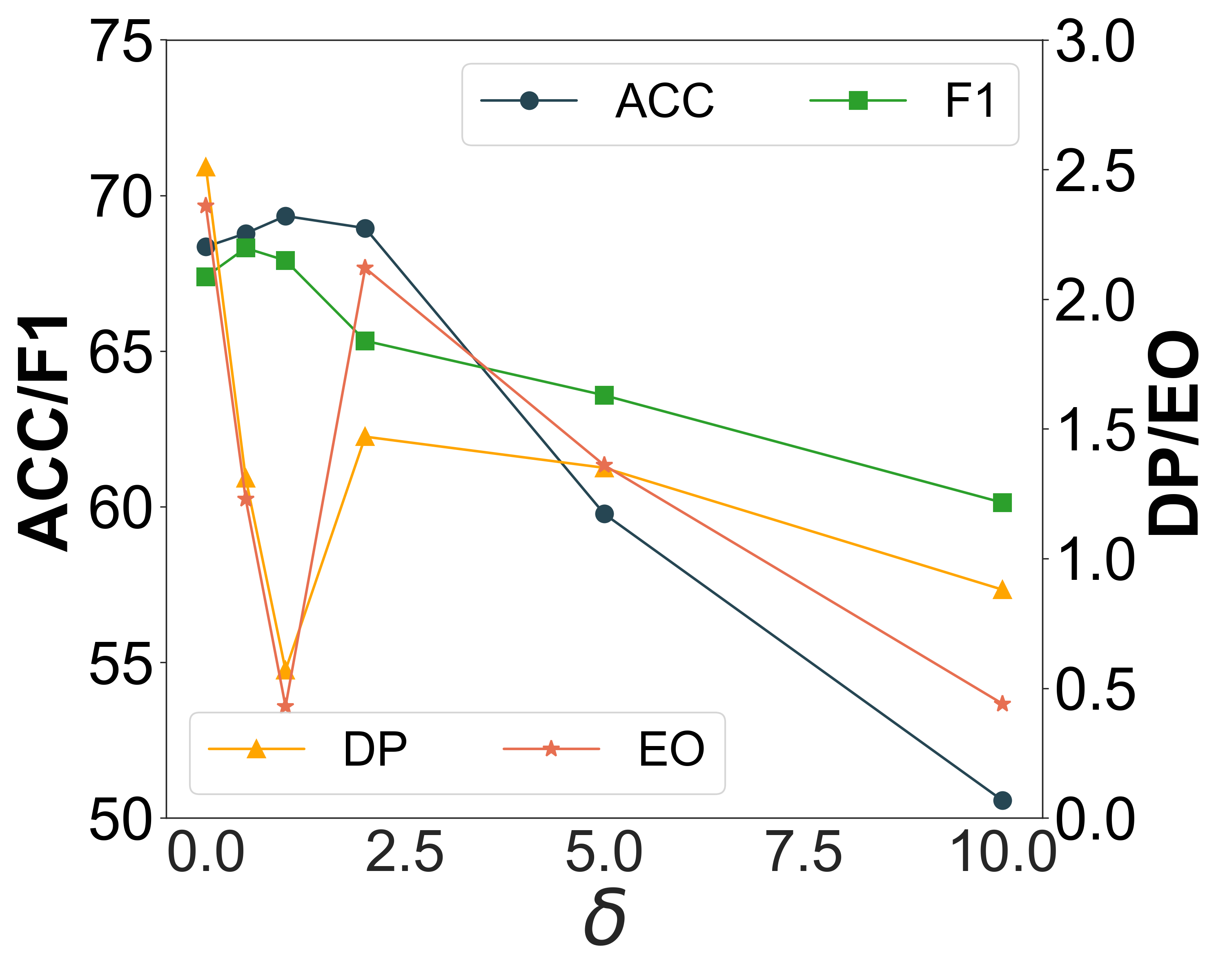}
  }
  \subfloat[{Pokec-z}]{
    \includegraphics[scale=0.15]{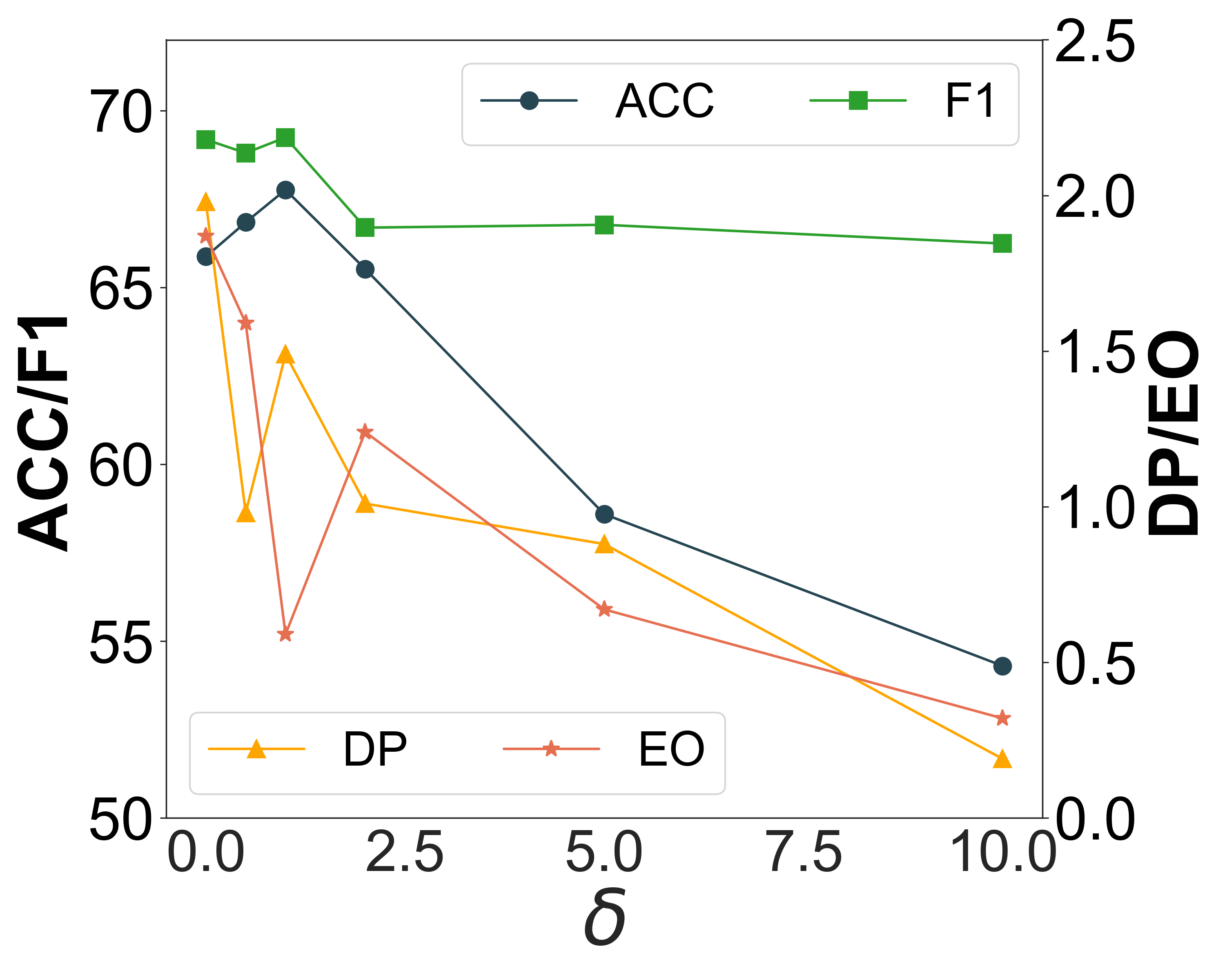}
  }
  \caption{Classification performance and group fairness under different values of hyper-parameter $\delta$.}
  \label{fig:hyper}
\end{figure}
\subsection{Hyper-parameter Analysis (RQ3)}
The value of $\delta$ is crucial to FairSIN as it can control the amount of introduced heterogeneous information. It is important to choose a proper value of $\delta$, as setting it too large may lead to sensitive information leakage in an opposite direction. We investigate the effect of hyper-parameter $\delta$ over \{0, 0.5, 1, 2, 5, 10\} with GCN encoder, and present the results in Figure~\ref{fig:hyper}. 
For Pokec-n and Pokec-z datasets, we can observe an optimal value $\delta=1$, where a favorable trade-off between predictive performance and fairness can be reached. The distribution of heterogeneous neighbors is too sparse on Pokec dataset as we can see in Table \ref{tab:dataset}, thus fairness are improving when $\delta$ increases to 10. In terms of predictive performance on Pokec-n, both accuracy and F1 score exhibit a decline as $\delta$ increases. As for Pokec-z, a similar trend is observed with the exception that the F1 score maintains a relatively stable level. In general, excessively large values of $\delta$ contribute to a decrease in predictive performance for both datasets. These observations align with our idea of \textit{neutralization}. Hyper-parameter experiment on German, Credit and Bail can be found in Appendix.
\begin{figure}[h]
  \centering
    \subfloat[{Bail}]{
    \includegraphics[scale=0.25]{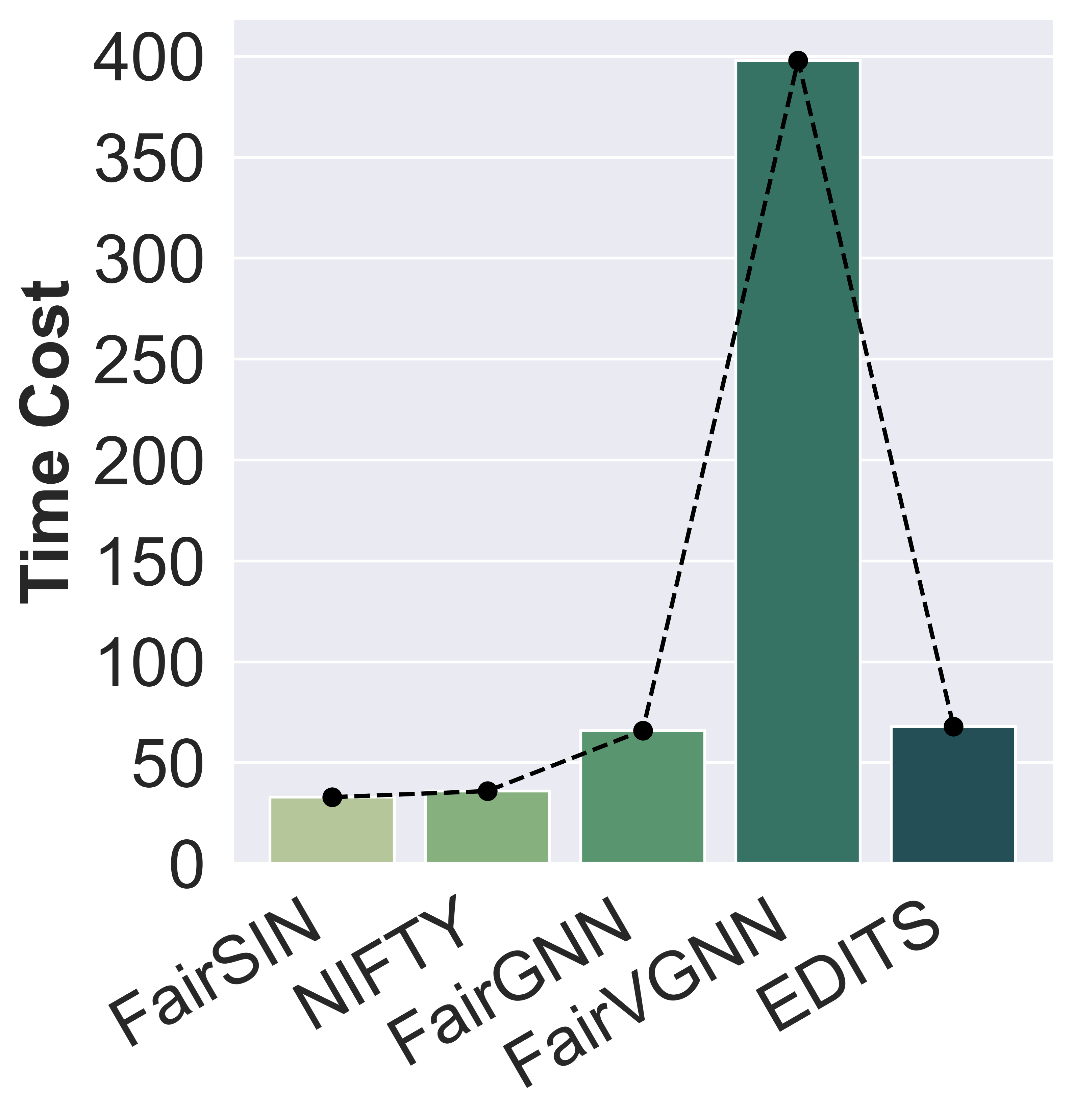}
  }
  \subfloat[{Credit}]{
    \includegraphics[scale=0.25]{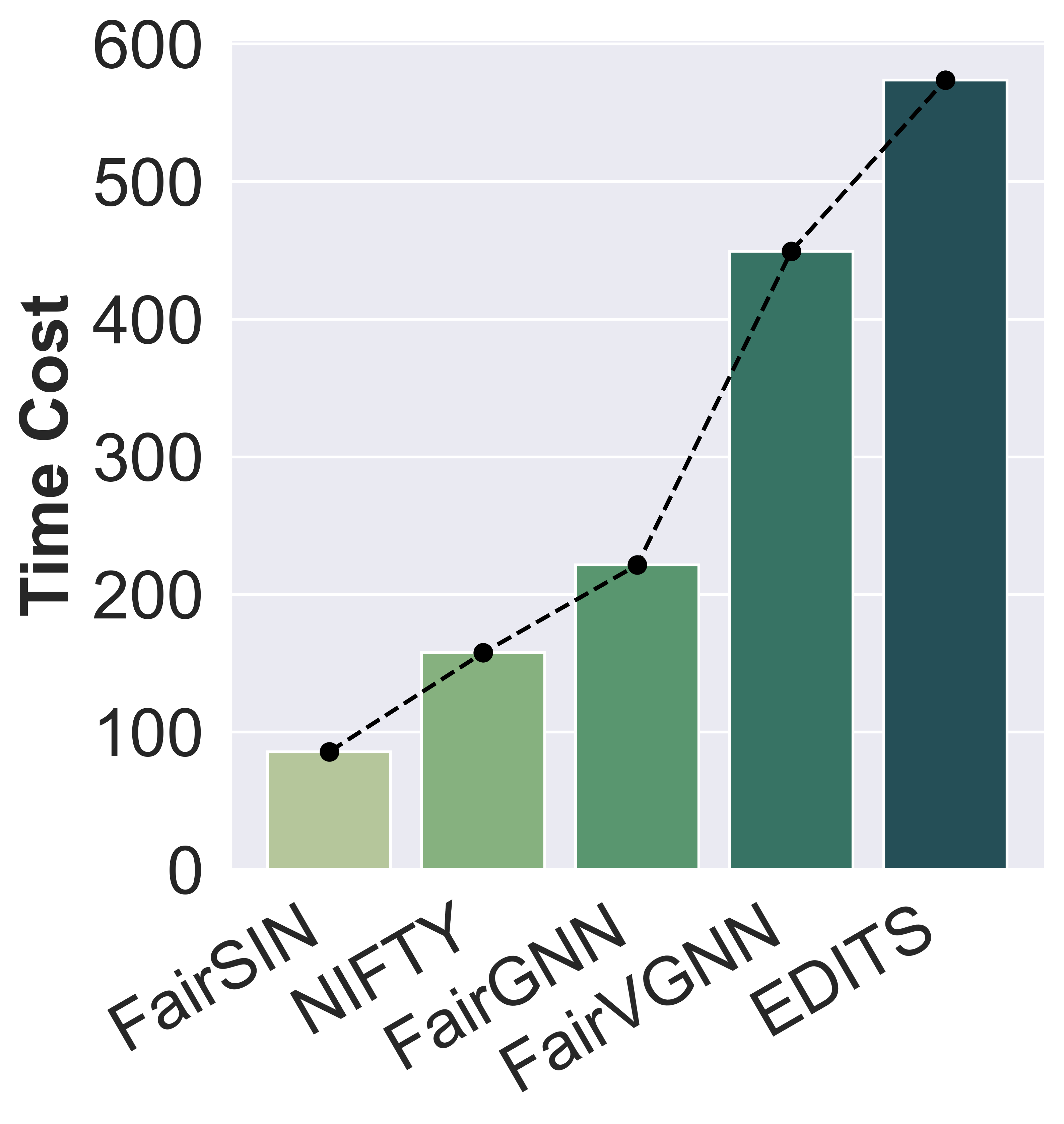}
  }
  \caption{Training time cost on Bail and Credit with GCN backbone (in seconds).}
  \label{fig:time_cost}
\end{figure}
\subsection{Efficiency Analysis (RQ4)}
As shown in Figure \ref{fig:time_cost}, we compare the training time cost of our FairSIN with the baselines on Bail and Credit datasets\footnote{Pokec datasets have larger scales, but EDITS run out of memory on them.}. We can find that FairSIN has the lowest time cost among all methods. Thus our method is both efficient and effective, enabling potential applications in various scenarios. FairVGNN \cite{fairv} incurs such high time cost attributed to its large number of parameters and the process of adverserial training. Also, EDITS~\cite{edits} needs to model node similarities between all node pairs for edge addition, and thus incurs a high time complexity. 
\section{Conclusion}

In this paper, we propose the \textit{neutralization-based} strategy FairSIN for learning fair GNNs, where extra \textit{F3} are added to node features or representations before message passing. By emphasizing the features of each node's heterogeneous neighbors, \textit{F3} can simultaneously neutralize the sensitive bias in node representations and provide extra non-sensitive feature information. We further present three implementation variants from both data-centric and model-centric perspectives. Extensive experimental results demonstrate the motivation and effectiveness of our proposed method. 


\section*{Acknowledgements}
This work is supported in part by the National Natural Science Foundation of China (No. U20B2045, 62192784, U22B2038, 62002029, 62172052, 62322203) and Young Elite Scientists Sponsorship Program (No. 2023QNRC001) by CAST.
\bibliography{aaai24.bib}

\end{document}